\documentclass[10pt,twocolumn,letterpaper]{article}

\usepackage[pagenumbers]{cvpr} 
\usepackage{graphicx}
\usepackage{multirow}
\usepackage[ruled,vlined,linesnumbered]{algorithm2e}
\usepackage{algorithmic}
\usepackage{booktabs}
\usepackage{graphicx}
\usepackage{amssymb}
\usepackage[table]{xcolor}
\definecolor{oursblue}{RGB}{230,241,255}
\definecolor{cvprblue}{rgb}{0.21,0.49,0.74}
\usepackage[pagebackref,breaklinks,colorlinks,allcolors=cvprblue]{hyperref}

\def\confName{CVPR}
\def\confYear{2027}

\title{What to Reinforce: Long-Horizon Value-Guided Preference Optimization for Test-Time Adaptation in Vision-Language Navigation}

\author{
Zihao Zhang\textsuperscript{1},
Zhoujie Shi\textsuperscript{1},
Aming Wu\textsuperscript{2},
Yang Li\textsuperscript{1},
Sijia Zhang\textsuperscript{1},
Yahong Han\textsuperscript{1}\\
\textsuperscript{1}School of Artificial Intelligence, College of Intelligence and Computing, Tianjin University, China\\
\textsuperscript{2}School of Computer Science and Information Engineering, Hefei University of Technology, China\\
{\tt\small zhangzihao2490@tju.edu.cn, zhoujieshi@tju.edu.cn, amwu@hfut.edu.cn}\\
{\tt\small liyang1389@tju.edu.cn, 3024244296@tju.edu.cn, yahong@tju.edu.cn}
}

\begin{document}
\maketitle
\begin{abstract}
Test-time adaptation (TTA) enables Vision-and-Language Navigation (VLN) agents to adapt to unseen environments using navigation experience. However, when supervision only indicates whether a navigation attempt succeeds or fails, it is difficult to determine which intermediate decisions should be reinforced or corrected. Successful trajectories may contain inefficient detours, while failed trajectories may include useful decisions. Existing methods primarily rely on local decision signals or episode-level outcomes, making it difficult to capture the long-horizon consequences of actions. To address this challenge, we propose Multi-Horizon Outcome-Consistent Value Adaptation (MOVA), which learns long-horizon navigation values to guide decision-level preference optimization. Multi-Horizon Value Learning leverages verified navigation experience to construct candidate-level supervision across multiple temporal horizons, enabling discriminative long-horizon value estimation without fine-grained action supervision. Building upon these values, Value-Guided Preference Adaptation evaluates the reliability of learned values and translates terminal feedback into decision-level preferences, enabling evidence-grounded hindsight corrections for successful trajectories and differentiated negative preferences for failed trajectories. 
Experiments on REVERIE, R2R, and R2R-CE demonstrate consistent improvements in navigation success and path efficiency.
\end{abstract}    

\vspace{-15pt}
\section{Introduction}
\label{sec:introduction}

Vision-and-Language Navigation (VLN) requires embodied agents to follow natural-language instructions and navigate through three-dimensional environments~\cite{anderson2018vln,guo2026awarevln,yao2025navmorph}. Despite recent advances~\cite{chen2022duet,wang2024goat}, navigation policies trained offline often struggle in unseen environments due to variations in visual appearance and spatial layouts. Test-time adaptation (TTA) addresses this challenge by updating navigation policies using test-time experience~\cite{gao2024fstta,kim2025feedtta}.

Navigation is inherently a long-horizon decision-making process~\cite{song2025longhorizon,wang2026progressthink}, where individual actions influence not only immediate observations but also future exploration and eventual task outcomes. Existing VLN adaptation methods exploit different forms of navigation feedback to guide policy updates. FeedTTA~\cite{kim2025feedtta} learns from binary episode-level outcomes, while ATENA~\cite{ko2025atena} selectively queries uncertain episodes and employs outcome-conditioned entropy optimization. To obtain finer-grained decision signals, FETA~\cite{wang2026feta} introduces action-level correctness queries, whereas CGPI~\cite{li2026cgpi} estimates decision contributions from local action-induced changes without external feedback. Despite these advances, trajectory-level outcomes cannot distinguish individual decision contributions, while local signals may overlook their long-horizon consequences. Under selective binary terminal feedback, extracting meaningful long-horizon decision information from limited navigation experience and translating it into effective policy optimization remains challenging.

\begin{figure}[t]
    \centering
    \includegraphics[width=\columnwidth]{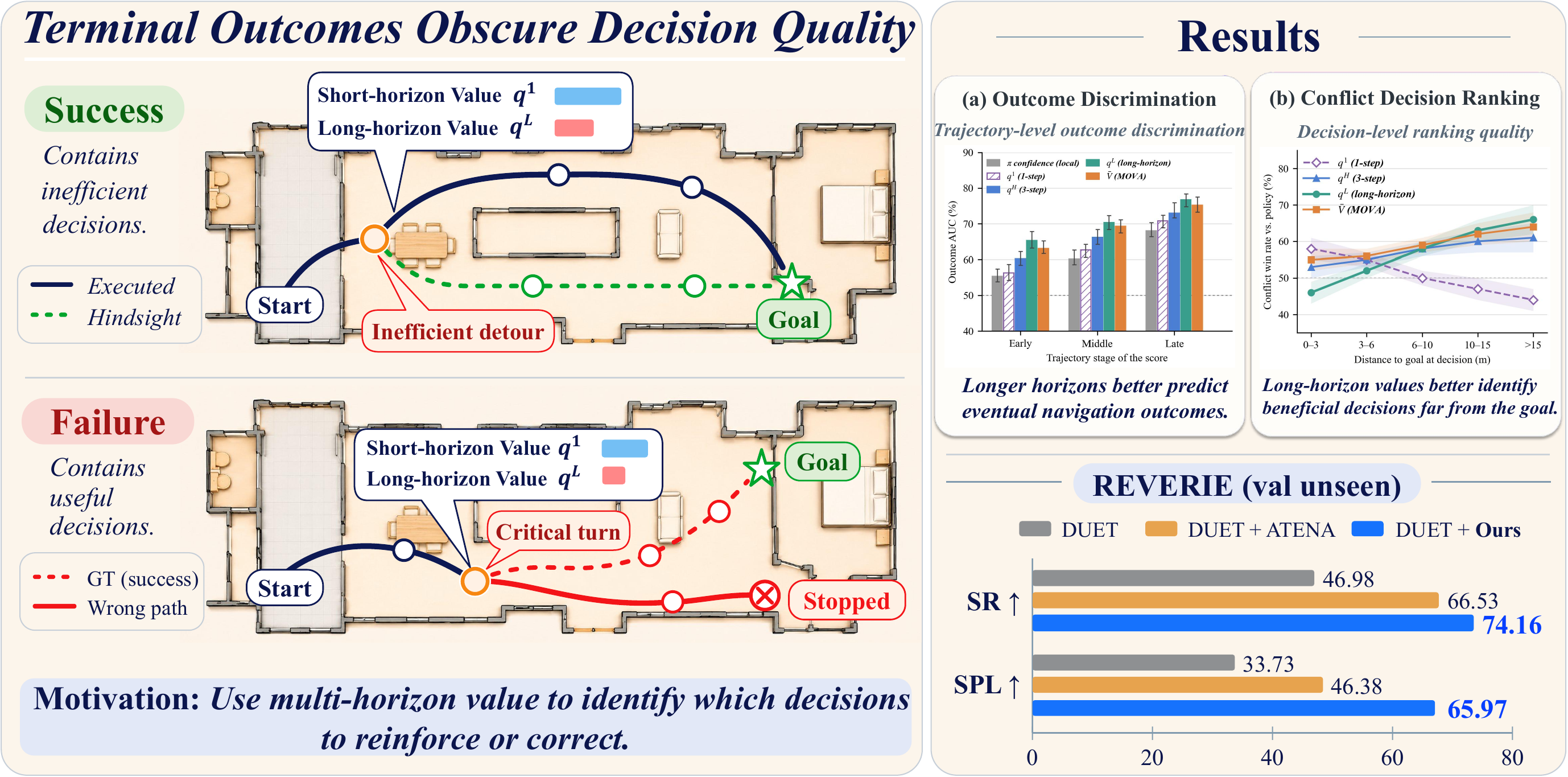}
    \vspace{-15pt}
    \caption{\textbf{Motivation and empirical evidence for MOVA.}
    \textbf{Left:} Successful trajectories may contain inefficient detours, while failed trajectories may still include beneficial decisions.
    \textbf{Right:} On REVERIE Val Unseen, (a) longer-horizon values exhibit stronger terminal outcome discrimination across trajectory stages; (b) they improve candidate decision ranking under policy disagreement, especially farther from the goal. The bottom panel compares navigation success and path efficiency.}
    \label{fig:motivation}
    \vspace{-15pt}
\end{figure}

The fundamental difficulty lies in the discrepancy between local decision quality and long-horizon navigation outcomes. As illustrated in Fig.~\ref{fig:motivation}(left), successful trajectories may contain inefficient detours, while failed trajectories may include beneficial decisions before critical mistakes. Thus, terminal outcomes alone cannot reliably identify which decisions should be reinforced or corrected. Fig.~\ref{fig:motivation}(right) provides further evidence on REVERIE Val Unseen: (a) outcome AUC evaluates how well different signals distinguish terminal successes from failures across navigation stages; (b) conflict win rate measures how effectively value estimates identify better candidate actions when disagreeing with the navigation policy, across different goal distances. Longer-horizon values exhibit stronger outcome discrimination and more reliable decision ranking, particularly farther from the goal. These findings raise two key challenges: (1) \textit{how to learn discriminative long-horizon navigation values} from sparse terminal feedback without intermediate supervision; and (2) \textit{how to construct reliable decision-level preferences} using learned values and verified navigation evidence for selective reinforcement and correction across successful and failed navigation trajectories.

To address these challenges, we propose Multi-Horizon Outcome-Consistent Value Adaptation (MOVA), which learns discriminative long-horizon navigation values and translates them into reliable decision-level preferences. MOVA comprises two complementary components: Multi-Horizon Value Learning and Value-Guided Preference Adaptation. The former estimates the long-horizon contributions of candidate decisions, while the latter uses reliable values and verified navigation evidence for selective reinforcement and correction. MOVA operates under selective binary terminal feedback without requiring stepwise rewards, expert actions, or privileged maps.

Specifically, Multi-Horizon Value Learning leverages verified successful trajectories to construct supervision for long-horizon value prediction, rather than treating all successful decisions as equally beneficial. Using verified endpoints and observed navigation topology, it derives candidate-level progress targets anchored to the same successful outcome. Observed navigation continuations extend supervision to alternative candidates, enabling their evaluation beyond immediate progress. Learning across multiple temporal horizons, together with candidate ranking constraints, enables discriminative estimation of long-horizon decision values. Failed and unqueried trajectories further support temporal consistency without introducing unavailable terminal labels, while auxiliary terminal outcome supervision encourages outcome-aware value representations.

Building upon these values, Value-Guided Preference Adaptation identifies which decisions should be reinforced or corrected under sparse terminal feedback. For successful trajectories, verified navigation evidence identifies shorter-route alternatives for hindsight correction, while historical ranking and outcome reliability guide the emphasis on corrected and retained decisions. For failed trajectories, reliable long-horizon values distinguish decision contributions and assign differentiated negative preference weights without assuming successful alternatives. Trajectory-wise normalization and reference-constrained preference optimization further enable selective policy adaptation.

Extensive experiments on REVERIE~\cite{qi2020reverie}, R2R~\cite{anderson2018vln}, and R2R-CE~\cite{krantz2020beyond} validate the effectiveness of MOVA, demonstrating consistent improvements in navigation success and path efficiency, as exemplified by the REVERIE Val Unseen results in Fig.~\ref{fig:motivation}(right).

Our main contributions are summarized as follows:

• We propose MOVA to address what to reinforce under sparse terminal feedback, transforming trajectory-level outcomes into reliable decision-level optimization signals.

• We develop outcome-consistent Multi-Horizon Value Learning and Value-Guided Preference Adaptation for evidence-grounded decision reinforcement and correction.

• Experiments on R2R, R2R-CE, and REVERIE demonstrate gains in navigation success and path efficiency.

\section{Related Work}
\label{sec:related_work}

\subsection{Test-Time Adaptation for VLN}
Test-time adaptation enables VLN agents to adapt using deployment experience. FSTTA~\cite{gao2024fstta} exploits unlabeled observations, while FeedTTA~\cite{kim2025feedtta} and ATENA~\cite{ko2025atena} leverage episodic feedback. FETA~\cite{wang2026feta} introduces action-level queries, whereas CGPI~\cite{li2026cgpi} estimates decision credit from local observations. However, terminal outcomes obscure individual decision contributions, while local signals may overlook long-horizon consequences. MOVA addresses these limitations by learning outcome-anchored long-horizon values to support more reliable and effective decision-level preference adaptation.

\subsection{Outcome-Guided Value Learning}
Prior studies in language reasoning explore outcome- and process-based supervision~\cite{uesato2022process,lightman2024verify}. OVM~\cite{yu2024ovm} estimates partial-path values from final outcomes, while subsequent methods investigate automated process supervision and long-horizon credit assignment~\cite{lee2025tvm,setlur2025rewarding,xu2025fullstepdpo,sullivan2026grpoprm,cheng2026graphgpo,auzina2026credit}. Unlike language reasoning, embodied navigation involves evolving observations and spatial dependencies, making intermediate supervision harder to construct. NavQ~\cite{xu2025navq} learns foresighted action representations from offline navigation trajectories, whereas MOVA learns outcome-anchored multi-horizon values from sparse test-time feedback to guide decision-level preference adaptation.

\begin{figure*}[t]
    \centering
    \includegraphics[width=\textwidth]{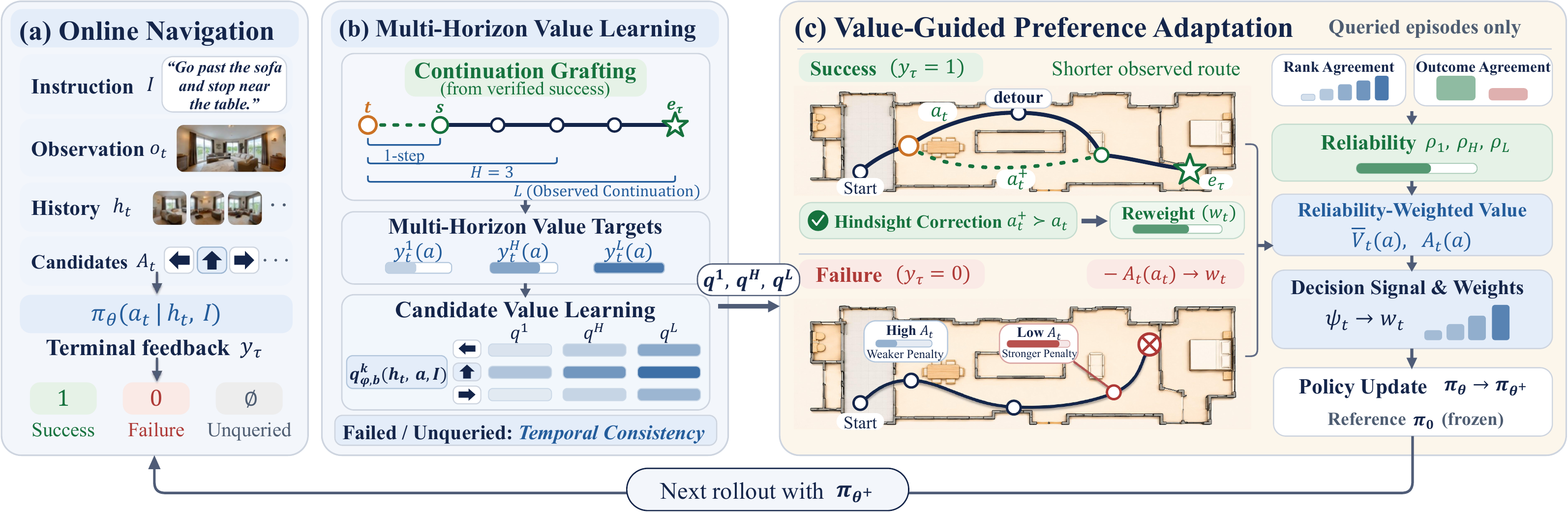}
    \vspace{-15pt}
    \caption{\textbf{Overview of MOVA.}
    \textbf{(a)} Online navigation with selectively queried binary terminal feedback.
    \textbf{(b)} Multi-Horizon Value Learning constructs success-anchored candidate targets through observed continuation grafting, while failed and unqueried episodes support temporal consistency.
    \textbf{(c)} Value-Guided Preference Adaptation uses historical ranking and outcome reliability to weight hindsight corrections on successes and differentiated negative preferences on failures, followed by reference-constrained policy updates.}
    \label{fig:mova_framework}
    \vspace{-10pt}
\end{figure*}

\section{Method}
\label{sec:method}
As illustrated in Fig.~\ref{fig:mova_framework}, MOVA transforms sparse terminal feedback into reliable decision-level signals through two components. Multi-Horizon Value Learning estimates outcome-anchored candidate values from observed continuations. Value-Guided Preference Adaptation leverages historical value reliability to guide hindsight corrections on successes and differentiated negative preferences on failures, enabling selective policy adaptation.

\subsection{Problem Definition}
\label{sec:problem}

During episode $\tau$, a VLN agent follows $\pi_{\theta}(a_t\mid h_t,I)$ for instruction $I$, where $h_t$ is its observation--action history and $a_t\in\mathcal A_t$ is a valid global navigation candidate or STOP. After executing $\tau$, the policy adapts as:
\begin{equation}
\theta^{+}=\mathcal U(\theta,\tau,y_\tau),
\label{eq:vln_tta}
\end{equation}
where $\mathcal U$ denotes the online update and $y_\tau\in\{1,0,\varnothing\}$ indicates success, failure, or unavailable feedback. Following ATENA~\cite{ko2025atena}, MOVA selectively queries binary terminal feedback after navigation episodes, without externally provided stepwise rewards or action-level supervision. Only queried episodes contribute to policy updates, while unqueried trajectories can support value learning without terminal labels. For a queried successful episode, its stopping viewpoint $e_\tau$ is confirmed as a valid endpoint by the terminal feedback. We maintain a frozen reference policy $\pi_0$, whose state and candidate embeddings provide a stable representation space for value learning.

\subsection{Multi-Horizon Value Learning}
\label{sec:mh_value}

\noindent\textbf{Success-Anchored Value Supervision.}
Rather than uniformly reinforcing successful decisions, we construct discriminative candidate-level value targets from verified successful episodes. Each episode builds a graph $G_\tau$ from observed navigable connections, recording their travel costs and directions without sharing edges across episodes. Given the verified successful endpoint $e_\tau$, let $d_\tau(v)$ denote the shortest distance from viewpoint $v$ to $e_\tau$ on $G_\tau$. At decision $t$, the agent is at $v_t$, and candidate $a$ leads to $v_t^a$ with travel cost $g_t(v_t,v_t^a)$ on the decision-time graph. We define:
\begin{equation}
\begin{aligned}
C_t(a)&=g_t(v_t,v_t^a)+d_\tau(v_t^a),\\
z_t(a)&=\operatorname{clip}_{[-1,0]}\!\left(
\frac{d_\tau(v_t)-C_t(a)}{\kappa}
\right),
\end{aligned}
\label{eq:route_efficiency}
\end{equation}
where $\kappa = 3\,\mathrm{m}$ controls the distance scale. The target $z_t(a)$ measures relative excess route cost, with zero indicating no observed detour. Since $d_\tau$ is computed retrospectively after successful feedback, endpoint information is not exposed during navigation.

Candidates with finite $g_t(v_t,v_t^a)$ and $d_\tau(v_t^a)$ are eligible for supervision. For visited destinations or $e_\tau$, $z_t(a)$ provides a point target; otherwise, it is treated as a censored lower bound $\underline z_t(a)$ to account for unexplored connections. Censored targets are supervised through a one-sided constraint, as detailed below. For STOP, we define the target across all horizons as:
\begin{equation}
z_t(\mathrm{STOP})
=\operatorname{clip}_{[-1,0]}
\left(-\frac{d_\tau(v_t)}{\kappa}\right),
\label{eq:stop_target}
\end{equation}
which is zero at the verified endpoint and negative elsewhere. These targets provide success-anchored supervision for learning candidate decision values.

\noindent\textbf{Candidate-Level Multi-Horizon Values.}
Building on the success-anchored targets, we use frozen $\pi_0$ embeddings to predict candidate values $q_{\phi,b}^{k}(h_t,a,I)\in(-1,0)$ for horizons $k\in\mathcal H=\{1,H,L\}$ and bootstrap members $b\in\{1,\ldots,B\}$. The three horizons represent the current decision ($1$), a fixed window of $H$ decisions, and the remaining observed continuation ($L$), respectively. Predictions are parameterized as $q_{\phi,b}^{k}=-\sigma(f_{\phi,b}^{k})$, where $f_{\phi,b}^{k}$ denotes the output logit.

After each episode, we construct candidate-level continuations from recorded navigation experience. For an eligible movement $a$ whose destination $v_t^a$ is subsequently visited at decision $s\leq t_e$, where $t_e$ denotes the first arrival at the verified endpoint, we define:
\begin{equation}
\mathcal Z_{t,a} =
\big(z_t(a),z_s(a_s),\ldots,
z_{t_e-1}(a_{t_e-1})\big).
\label{eq:observed_continuation}
\end{equation}
For executed actions, the suffix starts at the next decision; if $v_t^a=e_\tau$, the suffix is empty. This \emph{continuation grafting} combines a candidate's own target with the actually observed continuation from its destination, extending supervision beyond executed actions.

Let $z_j^{t,a}$ denote the $j$-th element of $\mathcal Z_{t,a}$, with $n_{t,a,H}=\min(H,|\mathcal Z_{t,a}|)$ and $n_{t,a,L}=|\mathcal Z_{t,a}|$. The multi-horizon targets are:
\begin{equation}
y_t^k(a)=
\frac{\sum_{j=0}^{n_{t,a,k}-1}\gamma^j z_j^{t,a}}
{\sum_{j=0}^{n_{t,a,k}-1}\gamma^j},
\quad k\in\{H,L\},
\label{eq:mh_target}
\end{equation}
where $\gamma$ is the discount factor and $n_{t,a,k}$ is the available continuation length. The one-step target is $y_t^1(a)=z_t(a)$. Shorter continuations are normalized by their actual lengths, and point-target candidates without a suffix reduce to one-step supervision. Censored frontiers retain only their one-step constraints, while STOP uses its endpoint-based target across all horizons. 
Unlike one-step targets, multi-horizon targets incorporate observed downstream decisions, distinguishing candidates with similar immediate progress but different long-term consequences. These values characterize outcome-anchored decision quality rather than calibrated success probabilities.
We set $H=3$ and $\gamma=0.8$.

\noindent\textbf{Multi-Horizon Value Learning Objectives.}
To mitigate value estimation instability under sparse terminal feedback, we adopt a bootstrap ensemble~\cite{osband2016deep} with $B=3$ independently initialized members. Each member is selected per decision with probability $p_{\mathrm{boot}}=0.8$; if none is selected, one is randomly assigned.

Let $\mathcal D_{p,b}$ collect valid targets selected for member $b$. For each sample $s\in\mathcal D_{p,b}$, we define $\ell_b(s)=(q_b^k-y_t^k)^2$ for point targets and $\ell_b(s)=[\underline z_t-q_b^1]_+^2$ for censored one-step targets. The value regression loss is:
\begin{equation}
\mathcal L_{p,b}
=\frac{\sum_{s\in\mathcal D_{p,b}}\ell_b(s)}
{\max(1,|\mathcal D_{p,b}|)},
\label{eq:progress_supervision}
\end{equation}
where $q_b^k=q_{\phi,b}^k(h_t,a,I)$. All valid targets are equally weighted, and empty sets contribute zero.

To enhance value discrimination, we construct same-state candidate ranking pairs separately for each horizon. For point targets, $a^+$ is preferred to $a^-$ when $y_t^k(a^+)-y_t^k(a^-)>\delta_r$. For $k=1$, a censored candidate may be preferred only if $\underline z_t(a^+)-y_t^1(a^-)>\delta_r$, with an exact target required for $a^-$. The ranking loss is:
\begin{equation}
\mathcal L_{r,b}=
\sum_{k\in\mathcal H}
\mathbb E_{\mathcal P_{k,b}}
\operatorname{softplus}\left(
-\frac{q_b^k(a^+)-q_b^k(a^-)}{T_r}
\right),
\label{eq:candidate_ranking}
\end{equation}
where $\mathcal P_{k,b}$ contains valid member-masked pairs, with shared margin $\delta_r=0.1$ and temperature $T_r=1$. This objective complements value regression with horizon-specific candidate comparisons.

We further introduce an auxiliary outcome head $o_{\phi,b}(h_t,a_t,I)$, supervised by queried terminal outcomes through $\mathcal L_{o,b}=\mathbb E_{\mathrm{qry},b}\operatorname{BCELogit}(o_{\phi,b},y_\tau)$. All executed decisions in a queried episode share its terminal outcome, providing auxiliary task-level supervision that encourages outcome-aware value representations.

For failed and unqueried trajectories, we impose temporal consistency using cached pre-update one-step predictions $\bar q_t^1=B^{-1}\sum_b\bar q_{t,b}^1(a_t)$. The consistency target is:
\begin{equation}
\bar y_{t,k}=
\frac{\sum_{j=0}^{n_{t,k}-1}\gamma^j
\operatorname{sg}(\bar q_{t+j}^1)}
{\sum_{j=0}^{n_{t,k}-1}\gamma^j},
\quad k\in\{H,L\},
\label{eq:cons_target}
\end{equation}
where $n_{t,H}$ covers at most $H$ recorded decisions, $n_{t,L}$ covers the remaining recorded suffix, and $\operatorname{sg}$ stops gradients. Recorded STOP actions are retained when present. All members minimize $\mathcal L_{c,b}=\mathbb E\sum_{k\in\{H,L\}}(q_b^k-\bar y_{t,k})^2$, maintaining multi-horizon consistency without additional terminal labels.

The overall value objective is:
\begin{equation}
\mathcal L_v=\frac1B\sum_{b=1}^B
\left(\mathcal L_{p,b}+\lambda_r\mathcal L_{r,b}
+\lambda_o\mathcal L_{o,b}+\lambda_c\mathcal L_{c,b}\right),
\label{eq:value_objective}
\end{equation}
where $\lambda_r=0.1$, $\lambda_o=1.0$, and $\lambda_c=0.5$. Bootstrap masks apply to labeled objectives, while temporal consistency exploits available failed and unqueried trajectories. 
The learned multi-horizon values subsequently guide reliability-weighted decision-level preference optimization.

\subsection{Value-Guided Preference Adaptation}
\label{sec:reliable_adaptation}

\noindent\textbf{Outcome-Grounded Decision Preferences.}
Successful and failed trajectories provide complementary decision-level evidence. For queried successes, let $t_e$ denote the first decision at the verified endpoint $e_\tau$. 
At each eligible decision, we select $a_t^+\ne a_t$ minimizing $C_t(a)$ and set $a_t^+\succ a_t$ if $C_t(a_t)-C_t(a_t^+)>\delta$ ($\delta=0.3\,\mathrm{m}$), ensuring corrections favor shorter routes on the observed graph.
Post-arrival movements are excluded from positive reinforcement. If the agent leaves $e_\tau$ at $t_e$, STOP is preferred when available; in continuous navigation, the movement at $t_e$ is also excluded and STOP corrections are disabled. For queried failures, executed decisions retain negative preferences without assuming successful alternatives, while reliable multi-horizon values differentiate their penalty strengths.

\noindent\textbf{Ranking and Outcome Reliability.}
We assess each value horizon through candidate ranking and terminal outcome agreement, using predictions cached before the corresponding episodes enter value training. Over the preceding $W$ queried episodes, $D_k^{\mathrm r}$ measures Somers' concordance on certified candidate pairs from earlier queried successes. Let $\mathcal T_\tau$ denote all recorded autonomous decisions and $\bar Q_\tau^k=|\mathcal T_\tau^{\mathrm o}|^{-1}\sum_{t\in\mathcal T_\tau^{\mathrm o}}\bar q_t^k(a_t)$ denote the mean cached value over decisions used for outcome evaluation. Here $\mathcal T_\tau^{\mathrm o}\subseteq\mathcal T_\tau$ includes STOP except in continuous navigation. We compute $D_k^{\mathrm o}$ as Somers' concordance between $\bar Q_\tau^k$ and queried episode outcomes. Both statistics measure concordant minus discordant pairs normalized by comparable pairs, with ties contributing zero to the numerator.

Let $N_{r,k}$ count certified ranking pairs for horizon $k$, and $N_o$ count successful--failed episode pairs. To ensure reliability, we set $\rho_k = 0$ if $N_{r,k} < N_r^{\min}$ or $N_o < N_o^{\min}$, with $N_r^{\min} = 1000$ and
$N_o^{\min} = 400$; otherwise, the joint reliability is:
\begin{equation}
\rho_k=\sqrt{[D_k^{\mathrm r}]_+[D_k^{\mathrm o}]_+},
\label{eq:joint_reliability}
\end{equation}
where $[x]_+=\max(x,0)$. Both statistics use earlier episodes only, and the geometric mean favors horizons consistent with both candidate rankings and terminal outcomes.

Let $R=\sum_{k\in\mathcal H}\rho_k$ and $\Lambda=R/|\mathcal H|$. For $R>0$, we aggregate cached multi-horizon values and define the candidate advantage:
\begin{equation}
\begin{aligned}
\bar V_t(a)
&=\frac{1}{R}\sum_{k\in\mathcal H}
\rho_k\bar q_t^k(a),\\
A_t(a)
&=\bar V_t(a)-
\sum_{c\in\mathcal A_t}
\pi_0(c\mid h_t,I)\bar V_t(c),
\end{aligned}
\label{eq:aggregated_value}
\end{equation}
where $\bar q_t^k(a)=B^{-1}\sum_b\bar q_{t,b}^k(a)$ is the cached ensemble prediction, and $\pi_0$ is renormalized over valid candidates, including STOP. Thus, $A_t(a)$ measures the decision value relative to reference-weighted alternatives. When $R=0$, value-guided reweighting is disabled.

\noindent\textbf{Reliability-Guided Preference Reweighting.}
Let $\mathcal T_{\rm hind}$ contain hindsight-corrected decisions and $\mathcal T_{\rm post}$ contain excluded post-arrival movements, both empty for failures. We define $\mathcal T_\tau^*$ as decisions eligible for reweighting, excluding $\mathcal T_{\rm post}$ and protocol-specific fixed-weight STOP decisions. For $R>0$, the decision-level value signal is:
\begin{equation}
\psi_t=
\begin{cases}
\bar V_t(a_t^+)-\bar V_t(a_t),
&t\in\mathcal T_{\rm hind},\\
A_t(a_t),
&y_\tau=1,\ t\notin\mathcal T_{\rm hind},\\
-A_t(a_t),
&y_\tau=0,
\end{cases}
\label{eq:decision_signal}
\end{equation}
where $\psi_t$ captures the relative value evidence supporting each preference. For successful episodes, larger value gaps emphasize hindsight corrections, while higher advantages prioritize retained decisions. For failed episodes, negated advantages strengthen penalties on relatively poor decisions while reducing penalties on potentially beneficial ones.

\begin{algorithm}[t]
\caption{Online Adaptation with MOVA}
\label{alg:mova}
\begin{algorithmic}[1]
\STATE Initialize $\pi_\theta$, frozen $\pi_0$, value ensemble $\phi$, replay $\mathcal D$, and reliability records $\mathcal P$.
\FOR{each instruction $I$}
    \STATE Compute $\{\rho_k\}$ from historical records $\mathcal P$ using Eq.~\eqref{eq:joint_reliability}.
    \STATE Execute $\tau$ with a fresh $G_\tau$; cache rollout data, pre-update predictions, and policy graph.
    \STATE Obtain selective terminal feedback $y_\tau\in\{1,0,\varnothing\}$.
    \IF{$y_\tau=1$}
        \STATE Construct success-anchored targets (Eqs.~\eqref{eq:route_efficiency}--\eqref{eq:mh_target}) and hindsight corrections.
    \ELSE
        \STATE Construct consistency targets (Eq.~\eqref{eq:cons_target}).
    \ENDIF
    \IF{$y_\tau\ne\varnothing$}
        \STATE Compute $\{w_t,m_t\}$ (Eqs.~\eqref{eq:normalized_weights}, \eqref{eq:refined_margin}).
        \STATE Update $\theta$ once using Eq.~\eqref{eq:policy_adaptation}.
    \ENDIF
    \STATE Add $\tau$ to $\mathcal D$; update $\phi$ using Eq.~\eqref{eq:value_objective}.
    \IF{$y_\tau\ne\varnothing$}
        \STATE Append cached reliability evidence to $\mathcal P$; retain the latest $W$ queried episodes.
    \ENDIF
\ENDFOR
\end{algorithmic}
\end{algorithm}

\begin{table*}[!t]
\centering
\vspace{-15pt}
\caption{
Results on the REVERIE dataset. Validation results are averaged over three seeds; $^{\dagger}$ denotes our re-implementation.
}
\vspace{-5pt}
\label{tab:reverie_results}
\setlength{\tabcolsep}{4pt}
\renewcommand{\arraystretch}{0.85}
\footnotesize
\resizebox{\textwidth}{!}{%
\begin{tabular}{l|cccc|cccc|cccc}
\hline
\multirow{2}{*}{\textbf{Method}} &
\multicolumn{4}{c|}{\textbf{Val seen}} &
\multicolumn{4}{c|}{\textbf{Val unseen}} &
\multicolumn{4}{c}{\textbf{Test unseen}} \\
& OSR$\uparrow$ & SR$\uparrow$ & SPL$\uparrow$ & RGSPL$\uparrow$
& OSR$\uparrow$ & SR$\uparrow$ & SPL$\uparrow$ & RGSPL$\uparrow$
& OSR$\uparrow$ & SR$\uparrow$ & SPL$\uparrow$ & RGSPL$\uparrow$ \\
\hline
HAMT~\cite{chen2021hamt}
& 47.65 & 43.29 & 40.19 & 25.18
& 36.84 & 32.95 & 30.20 & 17.28
& 33.41 & 30.40 & 26.67 & 13.08 \\
\hspace*{0.45em}+ FSTTA$^{\dagger}$~\cite{gao2024fstta}
& 47.62 & 43.27 & 40.15 & 25.14
& 36.88 & 32.90 & 30.14 & 17.20
& 33.39 & 30.39 & 26.65 & 13.61 \\
\hspace*{0.45em}+ IDEA~\cite{hu2026idea}
& 50.67 & 47.33 & 42.13 & 26.82
& 39.87 & 34.92 & 31.52 & 17.76
& 38.14 & 32.81 & 28.52 & 14.45 \\
\hspace*{0.45em}+ FeedTTA$^{\dagger}$~\cite{kim2025feedtta}
& 49.00 & 44.83 & 41.82 & 25.93
& 39.29 & 34.04 & 30.98 & 17.52
& 38.62 & 34.14 & 29.07 & 14.36 \\
\hspace*{0.45em}+ ATENA$^{\dagger}$~\cite{ko2025atena}
& 45.96 & 41.60 & 39.28 & 26.49
& 37.12 & 35.70 & 33.32 & 20.56
& 47.22 & 42.79 & 36.68 & 14.50 \\
\rowcolor{oursblue}
\hspace*{0.45em}\textbf{+ Ours}
& \textbf{68.40} & \textbf{61.91} & \textbf{58.60} & \textbf{36.86}
& \textbf{53.46} & \textbf{48.47} & \textbf{45.67} & \textbf{24.98}
& \textbf{59.94} & \textbf{54.20} & \textbf{49.28} & \textbf{18.96} \\
\hline
DUET~\cite{chen2022duet}
& 73.86 & 71.75 & 63.94 & 51.14
& 51.07 & 46.98 & 33.73 & 23.03
& 56.91 & 52.51 & 36.06 & 22.06 \\
\hspace*{0.45em}+ FSTTA$^{\dagger}$~\cite{gao2024fstta}
& 73.74 & 71.63 & 63.98 & 51.19
& 50.75 & 46.79 & 33.75 & 23.01
& 58.44 & 53.40 & 36.43 & 22.40 \\
\hspace*{0.45em}+ IDEA~\cite{hu2026idea}
& 78.45 & 78.24 & 67.74 & 55.07
& 58.51 & 56.92 & 38.03 & 25.47
& 58.91 & 55.12 & 39.84 & 24.52 \\
\hspace*{0.45em}+ FeedTTA$^{\dagger}$~\cite{kim2025feedtta}
& 78.45 & 77.42 & 71.02 & 57.92
& 64.03 & 59.18 & 39.37 & 26.34
& 58.76 & 53.58 & 37.66 & 24.10 \\
\hspace*{0.45em}+ FETA~\cite{wang2026feta}
& 84.31 & 82.29 & 73.75 & 59.01
& 73.27 & 68.33 & 50.55 & 33.15
& -- & -- & -- & -- \\
\hspace*{0.45em}+ ATENA$^{\dagger}$~\cite{ko2025atena}
& 83.65 & 82.24 & 73.40 & 58.76
& 69.86 & 66.53 & 46.38 & 30.91
& 57.74 & 54.28 & 40.70 & 25.01 \\
\rowcolor{oursblue}
\hspace*{0.45em}\textbf{+ Ours}
& \textbf{86.98} & \textbf{85.54} & \textbf{81.11} & \textbf{64.28}
& \textbf{77.30} & \textbf{74.16} & \textbf{65.97} & \textbf{43.56}
& \textbf{75.55} & \textbf{72.35} & \textbf{64.40} & \textbf{36.49} \\
\hline
GOAT~\cite{wang2024goat}
& 82.36 & 80.74 & 73.44 & 58.82
& 57.97 & 53.82 & 37.52 & 27.00
& 61.44 & 57.72 & 40.53 & 26.70 \\
\hspace*{0.45em}+ FSTTA$^{\dagger}$~\cite{gao2024fstta}
& 82.64 & 80.28 &72.81 & 57.90
& 57.94 & 53.79 & 37.49 & 26.98
& 62.35 & 57.52 & 39.49 & 26.82 \\
\hspace*{0.45em}+ FeedTTA$^{\dagger}$~\cite{kim2025feedtta}
& 87.07 & 85.85 & 79.30 & 64.37
& 68.44 & 65.52 & 46.68 & 33.22
& -- & -- & -- & -- \\
\hspace*{0.45em}+ FETA~\cite{wang2026feta}
& 90.02 & 87.65 & 77.98 & 62.14
& 72.75 & 68.65 & 54.22 & 38.18
& -- & -- & -- & -- \\
\hspace*{0.45em}+ ATENA$^{\dagger}$~\cite{ko2025atena}
& 87.94 & 86.77 & 79.15 & 63.75
& 72.64 & 69.75 & 49.42 & 34.91
& 64.26 & 62.03 & 46.82 & 31.54 \\
\rowcolor{oursblue}
\hspace*{0.45em}\textbf{+ Ours}
& \textbf{90.72} & \textbf{89.32} & \textbf{85.37} & \textbf{68.74}
& \textbf{79.31} & \textbf{76.82} & \textbf{68.86} & \textbf{48.56}
& \textbf{77.83} & \textbf{75.26} & \textbf{68.55} & \textbf{43.82} \\
\hline
\end{tabular}%
}
\vspace{-10pt}
\end{table*}

We compute $\widetilde w_t=1+\alpha\Lambda\tanh(\psi_t/\zeta)$ and normalize within each trajectory:
\begin{equation}
w_t=
\frac{|\mathcal T_\tau^*|\widetilde w_t}
{\sum_{j\in\mathcal T_\tau^*}\widetilde w_j},
\label{eq:normalized_weights}
\end{equation}
where $\alpha\in[0,1]$ controls reweighting strength and $\zeta>0$ controls sensitivity. When $R=0$ or $|\mathcal T_\tau^*|\leq1$, we set $w_t=1$; fixed-weight STOP decisions also retain unit weight. Normalization preserves a mean weight of one across eligible decisions while reallocating optimization emphasis according to reliable long-horizon values.

\noindent\textbf{Reference-Constrained Policy Optimization.}
Let $r_t(a)=\log\pi_\theta(a\mid h_t,I)-\log\pi_0(a\mid h_t,I)$ and $D_t=D_{\rm KL}(\pi_\theta\|\pi_0)$ over the recorded candidate set. With detached weights $w_t$, the decision-level preference margin is:
\begin{equation}
m_t=
\begin{cases}
0,
&t\in\mathcal T_{\rm post},\\
w_t[r_t(a_t^+)-r_t(a_t)],
&t\in\mathcal T_{\rm hind},\\
w_t[r_t(a_t)-\operatorname{sg}(D_t)],
&\text{otherwise},
\end{cases}
\label{eq:refined_margin}
\end{equation}
where $\operatorname{sg}$ stops gradients through the KL baseline. Hindsight corrections favor grounded alternatives, while retained decisions follow terminal outcome preferences.

Let $M_\tau=\sum_{t\in\mathcal T_\tau}m_t$, $s_\tau=2y_\tau-1\in\{-1,+1\}$, and $\bar D_\tau$ denote the mean KL divergence over all recorded decisions, including post-arrival steps. After each queried episode $\tau$, we perform one online policy update using:
\begin{equation}
\mathcal L_a(\tau)=
1-\sigma(\beta s_\tau M_\tau)
+\lambda_{\rm KL}\bar D_\tau,
\label{eq:policy_adaptation}
\end{equation}
where $\beta = 0.1$ controls preference sensitivity and
$\lambda_{\mathrm{KL}} = 0.02$ constrains deviation from $\pi_0$. Unlike conventional KTO~\cite{ethayarajh2024kto}, MOVA aggregates reliability-weighted decision margins within each episode before sigmoid optimization. Thus, successes reinforce retained or corrected decisions, while failures receive differentiated negative updates. The rollout policy remains fixed per episode, with cached reference and value predictions detached during adaptation.

\vspace{-5pt}
\subsection{Online Adaptation Procedure}
\label{sec:online_adaptation}

Algorithm~\ref{alg:mova} summarizes MOVA's online adaptation. Each episode constructs an independent observed topology without sharing graph edges. Queried successes provide success-anchored value targets and hindsight corrections, while failures contribute outcome supervision and temporal consistency. Unqueried trajectories support consistency learning without terminal labels. Reliability is estimated from preceding queried episodes, with current evidence recorded only after adaptation. Following the same entropy-based feedback query protocol as the baselines, MOVA updates the policy once per queried episode and continually trains the value model from recent replay. The reference policy remains frozen, while the adapted policy, value model, and replay persist across episodes.




\vspace{-5pt}
\section{Experiments}
\label{sec:experiments}

\subsection{Experimental Setup}

\noindent\textbf{Datasets and Metrics.}
We evaluate MOVA on VLN benchmarks covering discrete and continuous environments: REVERIE~\cite{qi2020reverie}, R2R~\cite{anderson2018vln}, and R2R-CE~\cite{krantz2020beyond}. REVERIE focuses on object-oriented navigation, R2R evaluates instruction following, and R2R-CE extends R2R to continuous environments. We report Val Seen and Val Unseen results for all benchmarks, and Test Unseen results for REVERIE. We use Success Rate (SR), Success weighted by Path Length (SPL), and benchmark-specific metrics.

\begin{figure*}[t]
    \centering
    \includegraphics[width=\textwidth]{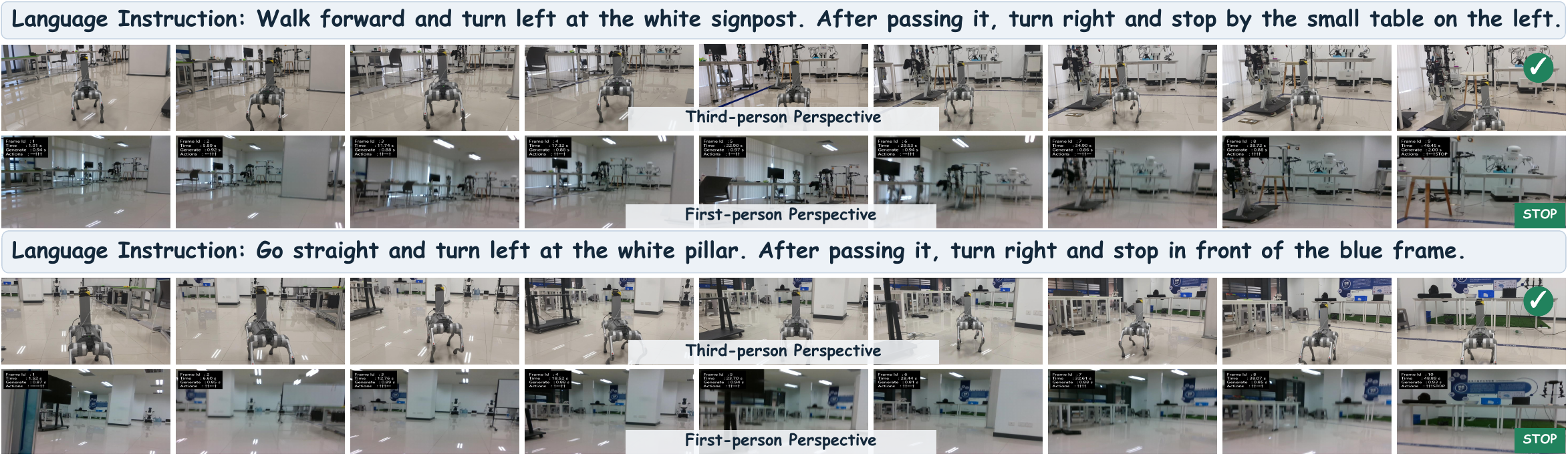}
    \vspace{-15pt}
    \caption{\textbf{Real-world qualitative results of MOVA on a Unitree Go2 robot.}
    Using an ETPNav~\cite{an2024etpnav} backbone adapted with MOVA, we show third-person views and egocentric observations from two language-guided navigation trajectories in indoor environments.}
    \vspace{-15pt}
    \label{fig:qual_real}
\end{figure*}

\noindent\textbf{Compared Methods and Protocol.}
We evaluate MOVA with five pretrained navigation backbones: HAMT~\cite{chen2021hamt}, DUET~\cite{chen2022duet}, GOAT~\cite{wang2024goat}, BEVBert~\cite{an2023bevbert}, and ETPNav~\cite{an2024etpnav}.
We compare against pretrained policies and representative TTA methods, including FSTTA~\cite{gao2024fstta}, FeedTTA~\cite{kim2025feedtta}, IDEA~\cite{hu2026idea}, ATENA~\cite{ko2025atena}, and FETA~\cite{wang2026feta}. Following the continual adaptation protocol, policies are updated sequentially using selective terminal feedback without resetting between episodes. Validation results are averaged over three seeds where applicable.

\noindent\textbf{Implementation Details.}
We follow a continual TTA protocol with uncertainty-based terminal feedback selection. Queried episodes provide binary outcomes for policy adaptation, while unqueried episodes support temporal consistency learning. MOVA employs a frozen reference policy and three bootstrap value models. Reliability is estimated over the latest $W=300$ queried episodes, with $\alpha=0.9$ and $\zeta=0.1$. All supervision uses episode-local observations without expert actions or privileged maps. 

\vspace{-5pt}
\subsection{Comparison with the State of the Art}

\begin{table}[t]
\centering
\caption{Results on the R2R dataset. Adaptation results are averaged over three seeds; $^{\dagger}$ denotes our re-implementation.}
\vspace{-5pt}
\label{tab:r2r_results}
\setlength{\tabcolsep}{3pt}
\renewcommand{\arraystretch}{0.96}
\resizebox{\columnwidth}{!}{%
\begin{tabular}{l|cccc|cccc}
\hline
\multirow{2}{*}{\textbf{Methods+Model}} &
\multicolumn{4}{c|}{\textbf{R2R Val seen}} &
\multicolumn{4}{c}{\textbf{R2R Val unseen}} \\
& TL$\downarrow$ & NE$\downarrow$ & SR$\uparrow$ & SPL$\uparrow$
& TL$\downarrow$ & NE$\downarrow$ & SR$\uparrow$ & SPL$\uparrow$ \\
\hline
DUET~\cite{chen2022duet}
& 12.33 & 2.28 & 79.00 & 73.00
& 13.94 & 3.31 & 72.00 & 60.00 \\
\hspace*{0.8em}+ FSTTA$^{\dagger}$~\cite{gao2024fstta}
&12.32 & 2.29 & 78.75 & 72.80 & 13.91 & 3.31 & 71.31 & 60.31 \\
\hspace*{0.8em}+ FeedTTA$^{\dagger}$~\cite{kim2025feedtta}
& \textbf{10.93} & 2.23 & 78.26 & 74.45
& 12.65 & 3.20 & 71.45 & 62.21 \\
\hspace*{0.8em}+ ATENA$^{\dagger}$~\cite{ko2025atena}
& 11.29 & 2.28 & 79.11 & 74.79
& \textbf{12.27} & 3.23 & 71.78 & 62.55 \\
\rowcolor{oursblue}
\hspace*{0.8em}\textbf{+ Ours}
& 11.76 & \textbf{2.09} & \textbf{80.67} & \textbf{75.85}
& 12.85 & \textbf{2.87} & \textbf{74.67} & \textbf{66.60} \\
\hline
BEVBert~\cite{an2023bevbert}
& 13.56 & \textbf{2.17} & 81.00 & 74.00
& 14.55 & 2.81 & 75.00 & 64.00 \\
\hspace*{0.8em}+ FSTTA$^{\dagger}$~\cite{gao2024fstta}
& 12.91 & 2.25 & 80.64 & 74.22 & 14.12 & 2.91 & 73.58 & 62.70 \\
\hspace*{0.8em}+ FeedTTA$^{\dagger}$~\cite{kim2025feedtta}
& \textbf{10.39} & 2.38 & 78.84 & 75.93
& \textbf{11.12} & 3.02 & 72.43 & 65.12 \\
\hspace*{0.8em}+ ATENA$^{\dagger}$~\cite{ko2025atena}
& 13.17 & \textbf{2.17} & 81.00 & 74.48
& 14.06 & 2.86 & 74.24 & 63.60 \\
\rowcolor{oursblue}
\hspace*{0.8em}\textbf{+ Ours}
& 11.65 & 2.19 & \textbf{81.42} & \textbf{77.55}
& 12.05 & \textbf{2.62} & \textbf{76.78} & \textbf{69.57} \\
\hline
GOAT~\cite{wang2024goat}
& 11.87 & 1.70 & 84.52 & 79.60
& 13.43 & 2.33 & 77.91 & 67.34 \\
\hspace*{0.8em}+ FSTTA$^{\dagger}$~\cite{gao2024fstta}
& 11.50 & 1.68 & 84.33 & 79.77 & 12.89 & 2.39 & 77.66 & 67.66 \\
\hspace*{0.8em}+ FeedTTA$^{\dagger}$~\cite{kim2025feedtta}
& \textbf{11.06} & 1.71 & 83.84 & 80.02
& 11.86 & 2.37 & 77.32 & 69.03 \\
\hspace*{0.8em}+ ATENA$^{\dagger}$~\cite{ko2025atena}
& 11.43 & 1.75 & 83.41 & 78.46
& 12.91 & 2.56 & 75.85 & 66.15 \\
\rowcolor{oursblue}
\hspace*{0.8em}\textbf{+ Ours}
& 11.39 & \textbf{1.65} & \textbf{84.72} & \textbf{80.63}
& \textbf{11.80} & \textbf{2.22} & \textbf{79.58} & \textbf{72.04} \\
\hline
\end{tabular}%
}
\vspace{-20pt}
\end{table}

\noindent\textbf{Evaluation on REVERIE.}
Table~\ref{tab:reverie_results} compares MOVA with existing TTA methods on REVERIE. MOVA consistently improves navigation success and path efficiency across HAMT, DUET, and GOAT. On Val Unseen, MOVA surpasses ATENA by \textbf{7.63} and \textbf{19.59} percentage points in SR and SPL with DUET, and by \textbf{7.07} and \textbf{19.44} points with GOAT, respectively.
These gains extend to Test Unseen, where GOAT+MOVA improves SR by \textbf{13.23} and SPL by \textbf{21.73} percentage points over ATENA, ranking \textbf{first in both SR and SPL} on the official REVERIE leaderboard\footnote{\url{https://yuankaiqi.github.io/REVERIE_Challenge/leaderboard2026.html}}. Re-implemented baselines follow their original formulations and reported configurations whenever applicable, with backbone-specific adjustments for stable adaptation. On REVERIE Test Unseen, both MOVA and our re-implemented ATENA obtain terminal feedback through a VLM gateway configured with the model identifier \texttt{gemini-4-argon}~\cite{google2026gemini4argon}.

\noindent\textbf{Evaluation on R2R and R2R-CE.}
Tables~\ref{tab:r2r_results} and~\ref{tab:r2rce_results} compare navigation results. On R2R Val Unseen, MOVA surpasses ATENA across three backbones by \textbf{3.05} and \textbf{5.30} percentage points in SR and SPL on average. On R2R-CE Val Unseen, MOVA improves SPL over ATENA by \textbf{3.53} and \textbf{3.48} percentage points with ETPNav and BEVBert, respectively. These results demonstrate MOVA's generalizability across architectures and environments.

\noindent\textbf{Real-world Qualitative Results.}
We deploy an ETPNav~\cite{an2024etpnav} policy adapted with MOVA in simulation on a real Unitree Go2 robot. Fig.~\ref{fig:qual_real} shows two language-guided navigation examples involving landmark recognition and directional changes. These results demonstrate the feasibility of MOVA for sim-to-real navigation transfer.

\begin{table}[t]
\centering
\caption{Results on the R2R-CE dataset. Adaptation results are averaged over three seeds; $^{\dagger}$ denotes our re-implementation.}
\vspace{-5pt}
\label{tab:r2rce_results}
\setlength{\tabcolsep}{3pt}
\renewcommand{\arraystretch}{0.96}
\resizebox{\columnwidth}{!}{%
\begin{tabular}{l|ccccc|ccccc}
\hline
\multirow{2}{*}{\textbf{Methods+Model}} &
\multicolumn{5}{c|}{\textbf{R2R-CE Val seen}} &
\multicolumn{5}{c}{\textbf{R2R-CE Val unseen}} \\
& TL$\downarrow$ & NE$\downarrow$ & OSR$\uparrow$ & SR$\uparrow$ & SPL$\uparrow$
& TL$\downarrow$ & NE$\downarrow$ & OSR$\uparrow$ & SR$\uparrow$ & SPL$\uparrow$ \\
\hline
ETPNav~\cite{an2024etpnav}
& 11.78 & 3.95 & 72.00 & 66.00 & 59.00
& 11.99 & 4.71 & 65.00 & 57.00 & 49.00 \\
\hspace*{0.8em}+ FSTTA$^{\dagger}$~\cite{gao2024fstta}
& 11.34 & 3.95 & \textbf{73.39} & 66.71 & 59.88
& \textbf{11.68} & 4.85 & 64.87 & 56.28 & 48.55 \\
\hspace*{0.8em}+ FeedTTA$^{\dagger}$~\cite{kim2025feedtta}
& \textbf{10.80} & 3.88 & 69.19 & 65.90 & 60.52
& 16.42 & 4.54 & 63.71 & 57.10 & 46.60 \\
\hspace*{0.8em}+ ATENA$^{\dagger}$~\cite{ko2025atena}
& 10.92 & 3.88 & 70.18 & 65.51 & 59.95
& 14.31 & 4.50 & \textbf{65.76} & 58.49 & 48.51 \\
\rowcolor{oursblue}
\hspace*{0.8em}\textbf{+ Ours}
& 11.45 & \textbf{3.81} & 72.24 & \textbf{67.99} & \textbf{61.79}
& 12.22 & \textbf{4.49} & 64.93 & \textbf{59.05} & \textbf{52.04} \\
\hline
BEVBert~\cite{an2023bevbert}
& 13.98 & 3.77 & 73.00 & 68.00 & 60.00
& 13.27 & 4.57 & 67.00 & 59.00 & 50.00 \\
\hspace*{0.8em}+ FSTTA$^{\dagger}$~\cite{gao2024fstta}
& 12.25 & 3.32 & 75.71 & 70.87 & 62.73
& 13.31 & 4.64 & 66.81 & 58.18 & 48.00 \\
\hspace*{0.8em}+ FeedTTA$^{\dagger}$~\cite{kim2025feedtta}
& 12.19 & \textbf{3.17} & \textbf{76.39} & \textbf{71.59} & 63.68
& 19.04 & 4.34 & 68.39 & 60.72 & 48.53 \\
\hspace*{0.8em}+ ATENA$^{\dagger}$~\cite{ko2025atena}
& \textbf{11.34} & 3.24 & 74.64 & 70.31 & 63.35
& 14.80 & 4.46 & 67.10 & 60.00 & 49.94 \\
\rowcolor{oursblue}
\hspace*{0.8em}\textbf{+ Ours}
& 11.81 & 3.20 & 74.81 & 70.57 & \textbf{63.86}
& \textbf{12.66} & \textbf{4.20} & \textbf{68.95} & \textbf{61.61} & \textbf{53.42} \\
\hline
\end{tabular}%
}
\vspace{-15pt}
\end{table}

\subsection{Ablation Study}
\noindent\textbf{Component Analysis.}
We conduct component ablations using DUET on REVERIE Val Unseen (Seed 1) to evaluate the contributions of MHVL and two key mechanisms in VGPA: hindsight correction and reliability-guided reweighting. As shown in Table~\ref{tab:component_analysis}, removing hindsight correction substantially degrades navigation performance, particularly SPL, highlighting its role in correcting suboptimal decisions within successful trajectories. Replacing MHVL with single-horizon value learning consistently reduces performance, demonstrating the importance of capturing long-term decision contributions. Disabling reweighting also leads to performance degradation, indicating that reliable value guidance enables more effective decision-level preference optimization. Together, these results demonstrate the complementary contributions of the proposed mechanisms.

\begin{table}[h]
\centering
\vspace{-5pt}
\caption{Component ablation of MOVA on REVERIE Val Unseen using DUET with seed 1.}
\vspace{-10pt}
\label{tab:component_analysis}
\setlength{\tabcolsep}{4pt}
\renewcommand{\arraystretch}{0.96}
\resizebox{\columnwidth}{!}{%
\begin{tabular}{l|c|cc|ccc}
\hline
\multirow{2}{*}{\textbf{Variants}} &
\multirow{2}{*}{\textbf{MHVL}} &
\multicolumn{2}{c|}{\textbf{VGPA}} &
\multirow{2}{*}{\textbf{SR}$\uparrow$} &
\multirow{2}{*}{\textbf{SPL}$\uparrow$} &
\multirow{2}{*}{\textbf{RGSPL}$\uparrow$} \\
& & \textbf{Hindsight} & \textbf{Reweighting}
& & & \\
\hline
KTO-base
& & & & 66.97 & 45.16 & 30.15 \\
w/o Hindsight
& \checkmark & & \checkmark & 69.94 & 57.96 & 37.50 \\
Single-Horizon
& & \checkmark & \checkmark & 70.58 & 60.91 & 41.65 \\
w/o Reweighting
& \checkmark & \checkmark & & 71.39 & 62.71 & 40.27 \\
\rowcolor{oursblue}
\textbf{Full MOVA}
& \checkmark & \checkmark & \checkmark
& \textbf{74.89} & \textbf{66.57} & \textbf{43.97} \\
\hline
\end{tabular}%
}
\vspace{-10pt}
\end{table}

\begin{figure}[t]
    \centering
    \includegraphics[width=\columnwidth]{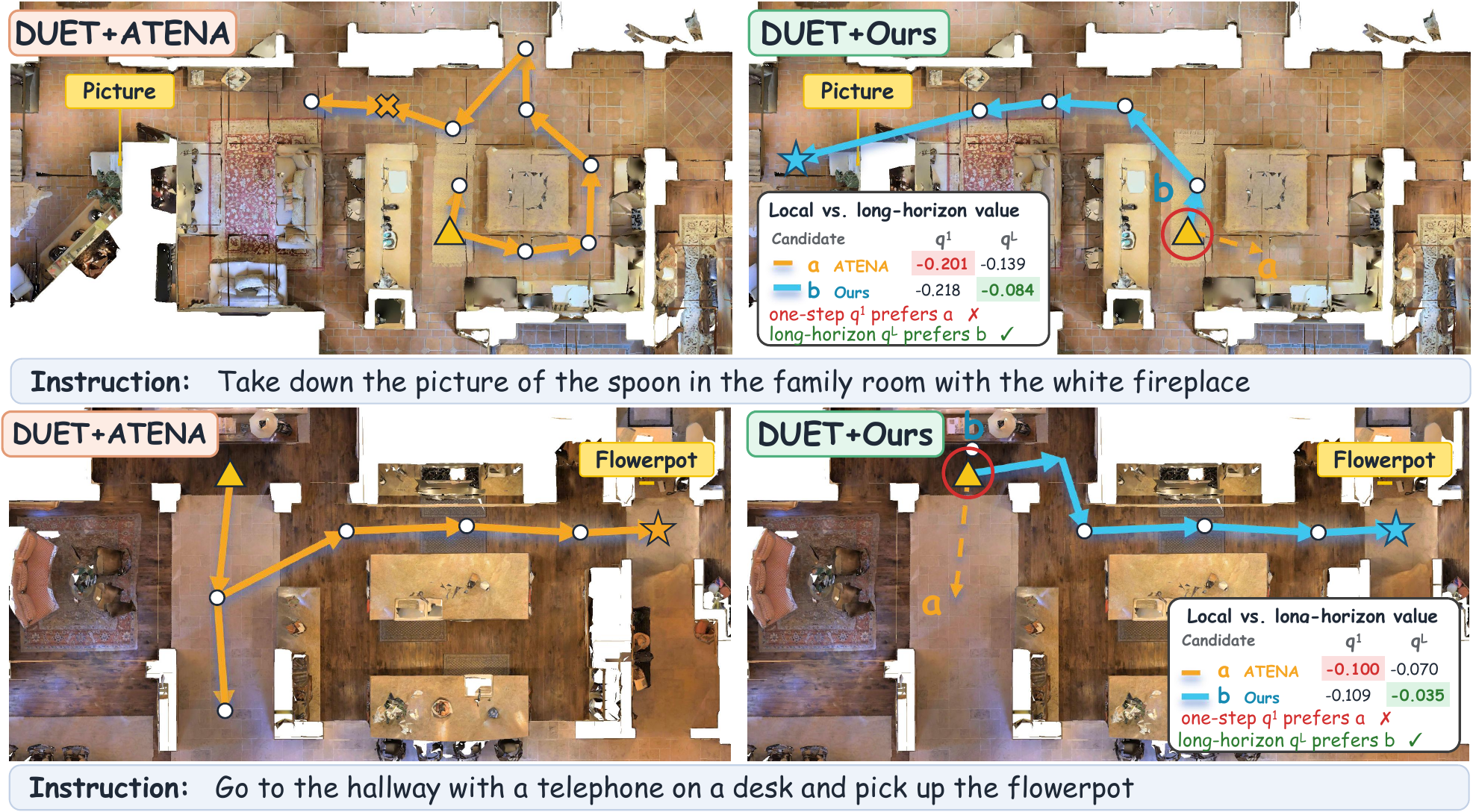}
    \vspace{-20pt}
    \caption{\textbf{Qualitative analysis on REVERIE.}
MOVA achieves more effective navigation than ATENA, with long-horizon values better distinguishing beneficial decisions than one-step values.}
\vspace{-5pt}
    \label{fig:qual_sim}
\end{figure}

\begin{figure}[h]
    \centering
    \includegraphics[width=\columnwidth]{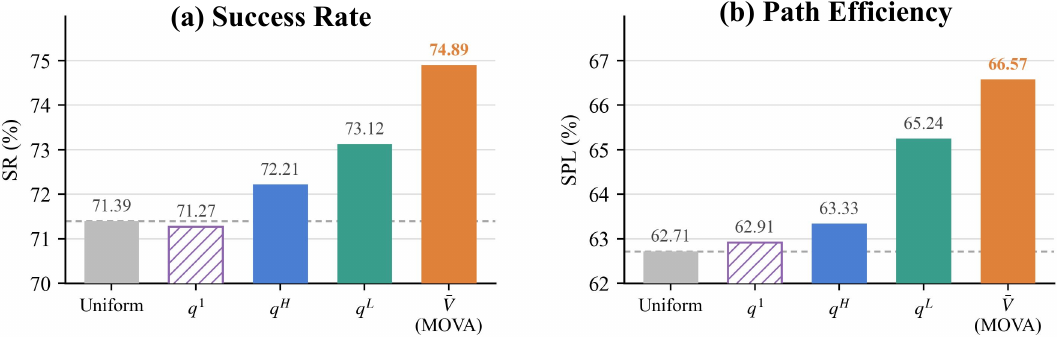}
    \vspace{-20pt}  
\caption{\textbf{Effectiveness of long-horizon value guidance.}
Comparison of uniform weighting, individual horizon values, and reliability-weighted aggregation on REVERIE Val Unseen using DUET, in terms of (a) SR and (b) SPL.}
\vspace{-10pt}
    \label{fig:value_analysis}
\end{figure}

\noindent\textbf{Effectiveness of Long-Horizon Value Guidance.}
We compare different value signals for decision-level preference adaptation on REVERIE Val Unseen. As shown in Fig.~\ref{fig:value_analysis}, one-step values perform comparably to uniform weighting, while longer-horizon values provide stronger guidance, improving both SR and SPL. Reliability-weighted multi-horizon aggregation further achieves the best performance, highlighting the benefits of combining complementary temporal information. Fig.~\ref{fig:qual_sim} provides qualitative evidence through two navigation examples, where long-horizon values favor more beneficial candidates than one-step values, corresponding to more successful and efficient navigation trajectories.

\begin{table}[h]
\centering
\vspace{-5pt}
\caption{Analysis of success and failure preferences on REVERIE Val Unseen using DUET with seed 1.}
\vspace{-10pt}
\label{tab:preference_analysis}
\setlength{\tabcolsep}{4pt}
\renewcommand{\arraystretch}{0.96}
\resizebox{\columnwidth}{!}{%
\begin{tabular}{l|cc|ccc}
\hline
\textbf{Variants} &
\textbf{Hindsight} &
\textbf{Failure RW} &
\textbf{SR}$\uparrow$ &
\textbf{SPL}$\uparrow$ &
\textbf{RGSPL}$\uparrow$ \\
\hline
w/o Both
& & & 67.68 & 48.03 & 31.95 \\
w/o Hindsight
& & \checkmark & 69.94 & 57.96 & 37.50 \\
w/o Failure RW
& \checkmark & & 71.82 & 63.45 & 42.53 \\
\rowcolor{oursblue}
\textbf{Full MOVA}
& \checkmark & \checkmark
& \textbf{74.89} & \textbf{66.57} & \textbf{43.97} \\
\hline
\end{tabular}%
}
\vspace{-10pt}
\end{table}

\noindent\textbf{Complementary Effects of Success and Failure Preferences.}
We further investigate the contributions of hindsight correction and failure reweighting on REVERIE Val Unseen. As shown in Table~\ref{tab:preference_analysis}, removing hindsight correction substantially reduces SPL, highlighting its role in refining suboptimal decisions within successful trajectories. Disabling failure reweighting also degrades navigation performance, indicating the importance of differentiating negative credit among failed decisions. Removing both mechanisms leads to the largest performance drop, demonstrating their complementary contributions to effective decision-level preference adaptation.

\begin{table}[t]
\centering
\caption{Analysis of ranking and outcome reliability on REVERIE Val Unseen using DUET with seed 1.}
\vspace{-5pt}
\label{tab:reliability_analysis}
\setlength{\tabcolsep}{4pt}
\renewcommand{\arraystretch}{0.96}
\resizebox{\columnwidth}{!}{%
\begin{tabular}{l|cc|ccc}
\hline
\textbf{Variants} &
\textbf{Ranking} &
\textbf{Outcome} &
\textbf{SR}$\uparrow$ &
\textbf{SPL}$\uparrow$ &
\textbf{RGSPL}$\uparrow$ \\
\hline
Uniform Reliability
& & & 72.33 & 64.36 & 41.87 \\
Ranking Only
& \checkmark & & 73.18 & 65.26 & 42.32 \\
Outcome Only
& & \checkmark & 73.01 & 65.09 & 43.23 \\
\rowcolor{oursblue}
\textbf{Joint (Ours)}
& \checkmark & \checkmark
& \textbf{74.89} & \textbf{66.57} & \textbf{43.97} \\
\hline
\end{tabular}%
}
\vspace{-15pt}
\end{table}

\noindent\textbf{Effectiveness of Joint Value Reliability.}
We examine the contributions of candidate ranking and terminal outcome agreement to value reliability estimation. As shown in Table~\ref{tab:reliability_analysis}, both signals improve navigation performance over uniform reliability weighting. Their combination achieves the best results across all metrics, demonstrating the complementary benefits of decision-level ranking and episode-level outcome evidence for preference adaptation.

\begin{figure}[h]
    \centering
     \vspace{-5pt}
    \includegraphics[width=1.0\columnwidth]{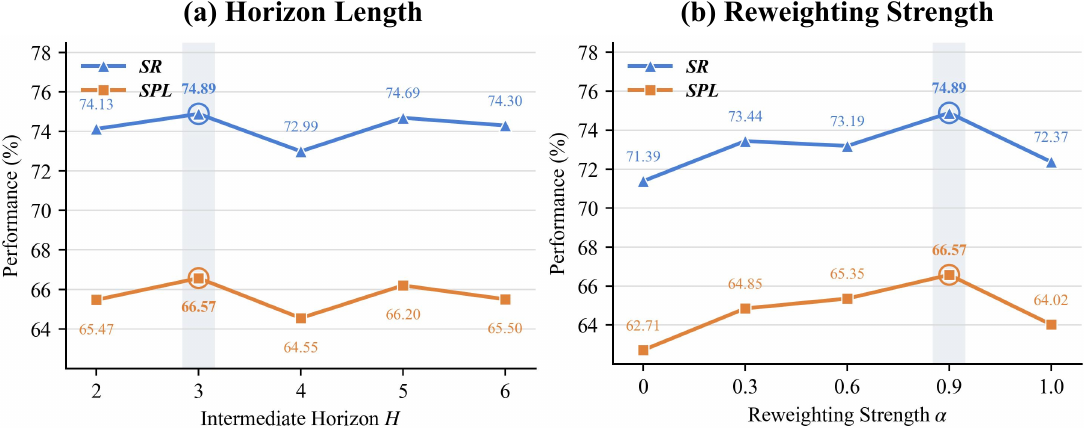}
    \vspace{-20pt}
    \caption{\textbf{Sensitivity analysis of MOVA with DUET on REVERIE Val Unseen.}
    Effects of (a) the intermediate horizon $H$ in MHVL and (b) the reweighting strength $\alpha$ in VGPA on navigation success rate (SR) and path efficiency (SPL).}
    \vspace{-5pt}
    \label{fig:hyperparameter_analysis}
\end{figure}

\noindent\textbf{Sensitivity to $H$ and $\alpha$.}
We examine the sensitivity of MOVA to the intermediate horizon $H$ in MHVL and reweighting strength $\alpha$ in VGPA. As shown in Fig.~\ref{fig:hyperparameter_analysis} (a), $H=3$ achieves the best performance, with other settings showing some fluctuations. Fig.~\ref{fig:hyperparameter_analysis} (b) shows that $\alpha=0.9$ yields the best results, while increasing it to 1.0 degrades performance, suggesting that excessive reweighting may amplify unreliable value guidance.

\vspace{-2pt}
\section{Conclusion}
\vspace{-2pt}
In this paper, we propose Multi-Horizon Outcome-Consistent Value Adaptation (MOVA) to address a fundamental challenge in test-time vision-and-language navigation: identifying which decisions to reinforce or correct under sparse terminal feedback. MOVA leverages verified successful trajectories to learn discriminative long-horizon navigation values, which guide reliable decision-level preference optimization. By combining evidence-grounded hindsight corrections for successful trajectories with differentiated negative preferences for failed trajectories, MOVA enables selective policy adaptation without intermediate rewards or expert supervision. 
Experiments on R2R, R2R-CE, and REVERIE demonstrate consistent gains in navigation success and path efficiency, validating MOVA's effectiveness in unseen navigation environments.


{
    \small
    \bibliographystyle{ieeenat_fullname}
    \bibliography{main}
}


\end{document}